\pdfoutput=1

\documentclass[11pt]{article}

\usepackage[]{ACL2023}

\usepackage{times}
\usepackage{latexsym,amssymb}
\usepackage{booktabs}
\usepackage{amsmath}
\usepackage{comment}
\usepackage{graphicx}
\usepackage{stmaryrd}
\usepackage{bm}
\usepackage{dsfont}
\usepackage{xspace}
\usepackage{longtable}
\usepackage{inconsolata}
\usepackage{algorithm2e}
\usepackage{multirow,multicol}
\usepackage{lscape}
\usepackage{tabularx}
\usepackage[capitalize]{cleveref}
	
\usepackage{soul}
\usepackage{pdflscape}
\usepackage{afterpage}

\DeclareMathOperator*{\argmax}{\arg\max}
\definecolor{red}{HTML}{EE220C}
\definecolor{blue}{HTML}{1434A4}
\definecolor{orange}{HTML}{ec5800}
\definecolor{green}{HTML}{006400}

\makeatletter
\crefname{section}{\S\@gobble}{\S\S\@gobble}
\crefname{subsection}{\S\@gobble}{\S\S\@gobble}
\makeatother

\usepackage[T1]{fontenc}

\usepackage[utf8]{inputenc}

\usepackage{microtype}

\usepackage{makecell}

\usepackage[disable]{todonotes}
\newcommand{\bzl}[2][]{\todo[color=red!25, #1]{\textbf{bzl}: #2}}
\newcommand{\jda}[2][]{\todo[color=blue!25, #1]{\textbf{jda}: #2}} 
 
\newcommand{\bzli}[2][]{\textcolor{red}{[\textbf{bzl}: #2]}}

\newcommand{\concat}{\cdot}

\title{Language Modeling with Latent Situations}

\author{
Belinda Z.\ Li ~~~~ Maxwell Nye ~~~~ Jacob Andreas \\ 
Massachusetts Institute of Technology \\
\texttt{\{bzl,mnye,jda\}@mit.edu}
}

\newcommand{\ourmethod}{\textsc{SituationSupervision}\xspace}
\newcommand{\ourmethodtable}{
\thead{\textsc{Situation-}\\\textsc{Supervision}}\xspace}

\begin{document}
\maketitle

\begin{abstract} Language models (LMs) often generate incoherent outputs: they refer to events and entity states that are incompatible with the state of the world
described in their inputs.  We introduce \ourmethod, a family of approaches for
improving coherence in LMs by training them to construct and condition on
explicit representations of entities and their states. \ourmethod has two components: an \textbf{auxiliary situation modeling} task
that trains models to predict state representations in context, and a
\textbf{latent state inference} procedure that imputes these states 
from partially annotated training data.
\ourmethod can be applied to both fine-tuning (by supervising LMs to encode state variables in their hidden representations) and prompting (by inducing LMs to interleave textual descriptions of entity states with output text).
In both cases, \ourmethod 
requires only a small number of
state annotations to produce major coherence improvements (between 4-11\%), showing that standard LMs can be sample-efficiently trained to model not just language but the situations it describes.
\end{abstract}

\begin{figure}[t!]
    \centering
    \includegraphics[width=0.95\columnwidth,trim={8cm 11cm 40cm 2cm},clip]{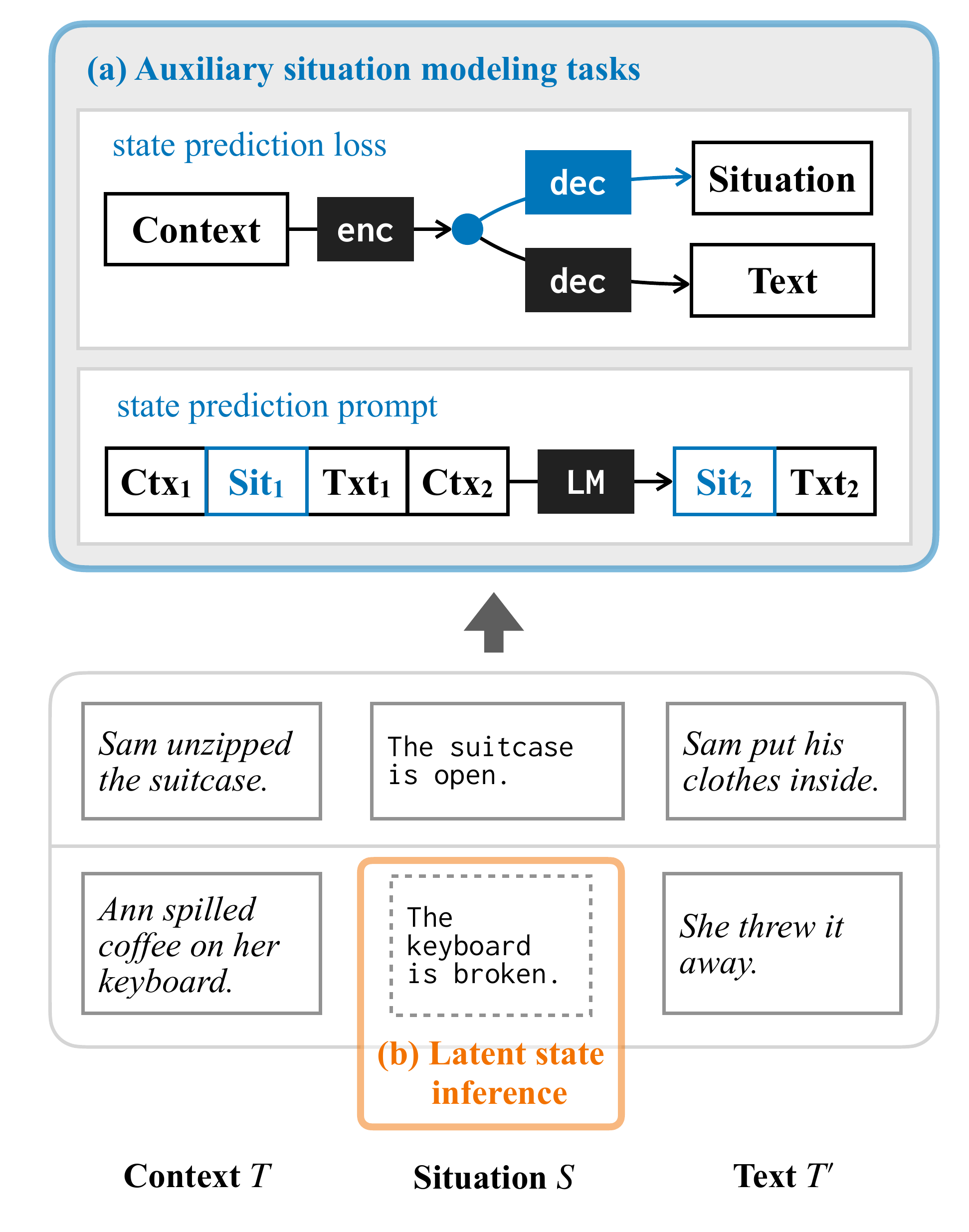}
    \caption{
    Language modeling with~\ourmethod, which comprises two components:
    (a) An auxiliary situation modeling task: given contexts annotated with explicit textual representations of the situations they describe, we use them to adapt LMs through either an auxiliary fine-tuning loss or a scratchpad-style prompt. %
    (b) A latent state inference procedure, whereby missing situation annotations are semi-supervisedly inferred, enabling auxiliary situation modeling starting from a small number of seed situation annotations.
    }
    \label{fig:teaser}
\end{figure}

\section{Introduction}

Recent years have seen dramatic improvements in the quality of text generated by
neural language models  \citep[LMs;][]{GPT3,T5}.  Nevertheless, even the best
LMs still suffer from \textbf{failures of semantic coherence}. Samples from LMs
refer to entities that have not yet been mentioned, assert contradictory facts,
or describe impossible sequences of events~\cite{marcus-and-davis}.  This paper introduces \ourmethod, a
family of methods for efficiently mitigating incoherent language generation by adapting
pre-trained LMs to \textbf{explicitly model the situations they describe} by tracking
the properties and relations of entities in generated text. The core of the proposed approach
is an auxiliary situation modeling task that trains LMs to predict textual
representations of entity state jointly with target text. However, for most
generation tasks,
state
information must be manually annotated and is costly to collect.
To make this auxiliary situation modeling task practical, we additionally introduce a
semi-supervised procedure for inferring entity states in unannotated text, making
it possible to perform auxiliary situation modeling with very small numbers of initial
state annotations.

Modern LMs can be specialized to new tasks in a variety of ways, including
fine-tuning their parameters and modifying their prompts. We develop versions of
\ourmethod suitable for both adaptation methods. For fine-tuned models, we
introduce an \emph{auxiliary state prediction loss} that encourages models' hidden
representations to encode state variables. For prompted models, we introduce a
\emph{scratchpad} approach that instructs models to generate explicit textual
descriptions of world states prior to generating output text. Both approaches
ultimately yield ordinary LMs, compatible with standard pre-training and
decoding procedures.\bzl{what does ordinary LMs mean? (in contrast to methods that don't yield ordinary LMs?)}

We evaluate \ourmethod on two challenging text generation tasks: the
TextWorld (TW) task of generating acceptable next actions in a text-adventure game~\citep{ct2018textworld}, and the TRIP task of evaluating commonsense physical plausibility of short (5-sentence) stories~\cite{storks-etal-2021-tiered-reasoning}. In experiments on
fine-tuned BART LMs~\cite{lewis-etal-2020-bart}, applying \ourmethod with 500 seed state annotations reduces coherence errors by 4\% on TW %
and 7\% on TRIP. In experiments on prompted GPT-3 models~\cite{GPT3}, %
12 seed state annotations reduce coherence errors by 8.2\% on TW
and 20 seed state annotations reduce errors by 7.6\% on TRIP.
In both cases, it is far more sample-efficient to provide state annotations for existing training samples than to augment training data with additional text-only samples: in fine-tuned models, \ourmethod with 500 state annotations can perform comparably to training on 9000 more text-only sentences, while in prompted models, %
devoting a fixed token budget to state annotations rather than additional text samples yields a coherence improvement of up to 10 points.

Additional experiments characterize the ingredients of a good  state
representation, showing that training LMs to predict \emph{causally relevant}
state variables is essential for good performance. Because the latent state
inference objective favors state representations that improve LM predictions,
\ourmethod discovers these variables automatically, sometimes improving on
human-designed state representations.
In summary:
\begin{enumerate}
  \item We show that training LMs to build explicit representations of entity
    state (via auxiliary losses or scratchpad-based prompting) improves
    coherence in text generation tasks.

  \item We describe new algorithms for \emph{semi-supervised} learning of state
    representations, enabling auxiliary supervision and scratchpad techniques to
    be applied with extremely small numbers of annotations.

\end{enumerate}
Our results show that, even when LMs struggle to generate coherent text, only a
small amount of annotation is needed to train them to infer and predict explicit
representations of situations. Once they can be predicted, these representations in turn
confer large improvements in LM coherence itself.

\section{Background and Related Work}

A \textbf{language model} (LM) encodes a distribution $p(T' \mid T)$ over texts
$T'$ given contexts $T$ (\cref{fig:teaser}).
Today, most LMs are implemented as deep neural networks trained on massive text
corpora~\cite{GPT3}.  Sampling from them produces naturalistic text that often
resembles human-generated language.  However, LM generation is prone to
several failure modes, including generation of text that is incoherent,
untruthful, or
unreliable~\cite{zhou-etal-2021-detecting,maynez-etal-2020-faithfulness,martindale-etal-2019-identifying}.
Past work has shown that some of these behaviors stem from models' failure to build
good representations, both of entities' default properties \citep{onoe2021creak} and state
changes in context \citep{zellers-etal-2021-piglet}. Humans' ability to avoid these failure modes, and to
generate truthful and coherent text, is generally understood to rest upon
explicit mental representations of the \emph{situations} that language
communicates. The nature and structure of these representations remains an
ongoing topic of research in linguistics and cognitive science, but
existing theories broadly agree that language users maintain explicit beliefs
about the properties of and relations among entities mentioned in a discourse, updating these
beliefs in response to new observations or new information conveyed in language \citep[e.g.][]{kratzer2007situations, zwaan2012revisiting}.

\begin{figure*}
    \centering
    \includegraphics[width=\textwidth,trim={0 1.5cm 0 0.5cm},clip]{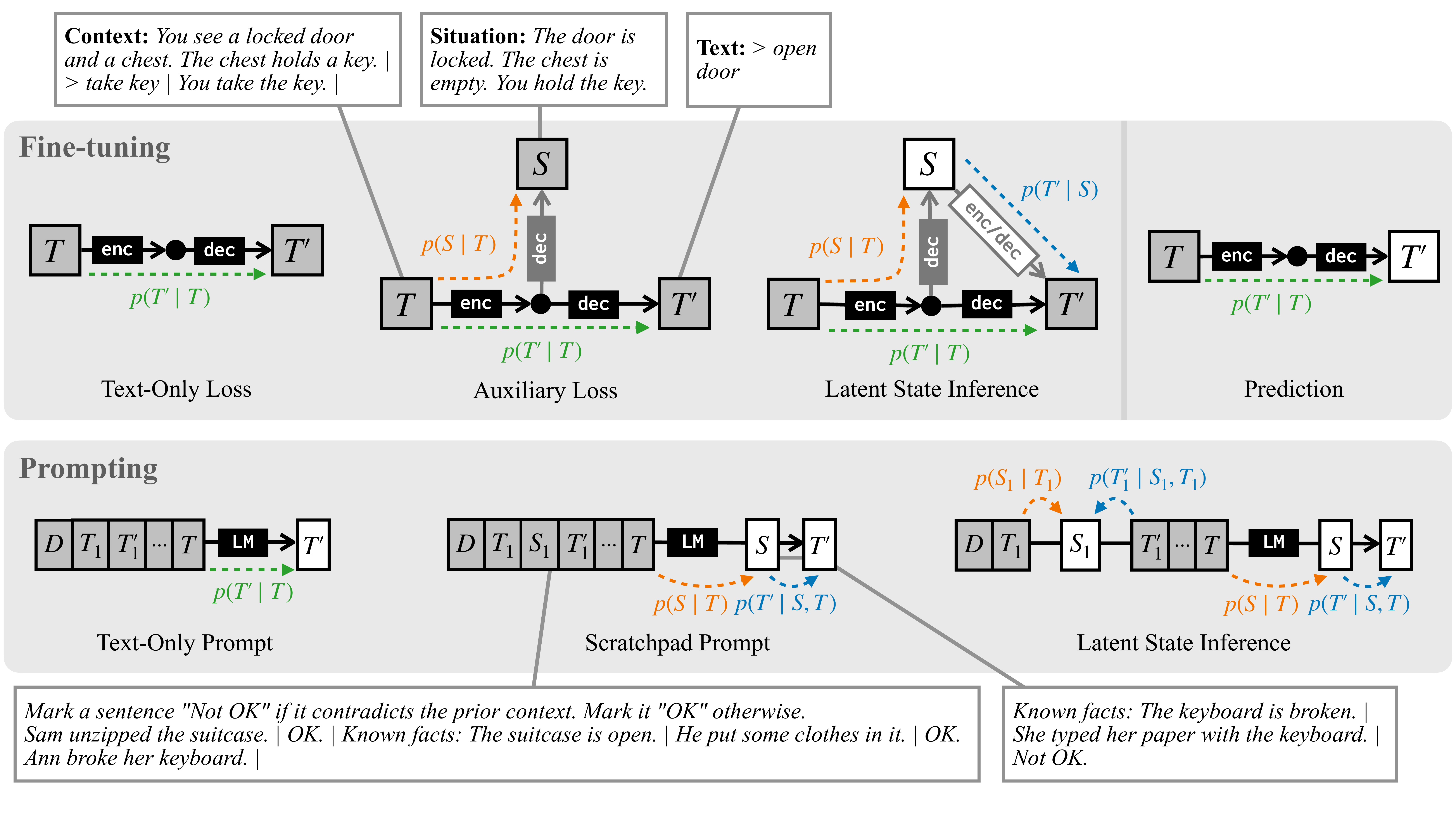}
    \caption{Fine-tuning (top) and prompting (bottom) with~\ourmethod. In each, we show the: \\
    \textit{Left}: normal text-only procedure, where the LM is trained/prompted with text only samples and expected to produce next sentences $T'$ from contexts $T$. \textit{Middle}: auxiliary situation modeling component of~\ourmethod, where the LM is given state descriptions $S$ during training or in the prompt and expected to learn to encode it in its parameters or infer it in-context. \textit{Right}: latent state inference component of~\ourmethod, where the LM must infer missing state descriptions in the training data or prompt demonstrations. \\
    Finally, in fine-tuning, state reasoning is done implicitly during inference-time, meaning the inference procedure is the same for all forms of training: we use the base LM to infer the next sentence from the context.
    }
    \label{fig:methods}
\end{figure*}

These representational theories suggest that language models $p(T' \mid T)$
might also benefit from explicit modeling of situation state.  Given an input text
$T$, such a model would begin by representing the \textbf{situation} $S$ described by $T$. Following \citet{barwise1981situations}, the situations we consider in this paper are not complete descriptions of possible worlds, but instead just sets of facts that are known or inferable about salient entities in a discourse.
Examples, with facts expressed as sentences in natural language, are shown in \cref{fig:teaser} and \cref{fig:methods}. Having inferred $S$ from $T$, a language model may condition on it when sampling $T'$ from $p(T' \mid S, T)$.

Past work
has proposed a number of language generation models that explicitly model the state of the world,
primarily by developing specialized prediction architectures that maintain
internal state representations \citep{henaff2016tracking,gupta2019tracking} or interact with outside
simulation engines \citep{liu2022mind}. While effective, these approaches come at a cost---requiring 
complex training data \citep{mishra2018tracking}, limiting models to narrow, pre-defined
domains, and generally precluding the large-scale (text-only) pretraining
responsible for many of the greatest successes of current LMs. The main question
this paper seeks to answer is whether the benefits of explicit world modeling
may be obtained entirely within the language modeling paradigm itself, without
the need for specialized model architectures or large amounts of specialized
supervision.

We do so by adapting pre-trained LMs to better represent
situations $S$. There are two standard frameworks for LM adaptation. In
smaller models, which are generally adapted by \textbf{fine-tuning} of model parameters, we develop
auxiliary loss functions that encourage models' hidden states to contain
the information required to generate textual descriptions of state.  In larger
models, which can also be \textbf{prompted} by pre-pending a task description or set of examples to the input context, we develop prompts
that induce models to generate textual state descriptions in LM output
itself. Our work builds on a large body of work that uses auxiliary prediction
tasks to shape model representations, notably work using ``scaffold'' decoders
to shape model representations of syntax \citep{swayamdipta2018syntactic,wilcox2019structural}, and and ``scratchpad'' or
``chain-of-thought'' approaches to perform intermediate computations in models'
output spaces \citep{camburu2018snli,nye2021show,wei2022chain}.  In \cref{sec:aux_supervision}, we show how to adapt both techniques for a
new class of open-ended text generation problems.

Adapting LMs with auxiliary prediction tasks requires a source of data for
auxiliary supervision. This kind of supervision is
uniquely difficult to obtain for open-ended generation tasks. But the
probabilistic framing described above makes it natural to formulate language
modeling with explicit situations as a latent variable problem. At training time, we
may use context $T$ and targets $T'$ to guide inference of the unknown $S$ from
which $T'$ was predicted. Once inferred, these states supervise the
representation-building model that predicts $S$ from $T$ alone. As above, a
great deal of past work has focued on treating string-valued prompts or plans as
latent variables \citep{sharma2021skill,zelikman2022star,sun2022black}. In \cref{sec:latent_inf}, we generalize these methods to
support multi-step, open-ended text generation, and show that inferred states
can be used to supervise small models as well as prompt large ones.

\section{Auxiliary Situation Modeling}
\label{sec:aux_supervision}

We begin by assuming access to a pre-trained LM and two sources of supervision:
a dataset $\mathcal{X}_U$ of text examples of the form $(T, T')$, and a smaller
dataset $\mathcal{X}_A$ of examples $(T, S, T')$ annotated with textual situation
descriptions $S$. Our full training data $\mathcal{X}$ is thus $\mathcal{X}_U\cup \mathcal{X}_A$. As depicted in \cref{fig:methods}, we take these state descriptions to consist of declarative sentences about the properties and relations of entities that are relevant to the text being generated.
In this section, we describe two auxiliary prediction
schemes that use these annotations to improve the LM's ability to model the
conditional text distribution $p(T' \mid T)$.

\subsection{Situation Modeling for Fine-tuning}
\label{sec:aux_method}
Our first approach uses a \emph{auxiliary decoding loss} that encourages
context representations to directly encode state information.  We focus on
encoder--decoder models consisting of an
encoder $\mathcal{E}$ and a decoder %
$\mathcal{D}$, with $\mathcal{D}(\mathcal{E}(T))$ producing as output a
probability distribution over next sentences $T'$. In standard training,
parameters of $\mathcal{E}$ and $\mathcal{D}$ are chosen to maximize:
\begin{align}
    \mathbin{\color{green}\mathcal{L}(T'|T)} =
    \log \mathbin{\color{green}p(T'|T)} = 
    \log \mathcal{D}(T' \mid \mathcal{E}(T))
\end{align}
To improve state representations, we add an \textbf{auxiliary loss}.
This takes the form of an auxiliary decoder 
$\mathcal{D}_{S|T}$ 
(distinct from the original decoder) which is trained to predict state
representations $S$ from the encoded context $\mathcal{E}(T)$. We define
\begin{align}
    \mathbin{\color{orange}\mathcal{L}(S|T)} =
    \log \mathbin{\color{orange}p(S|T)} =
    \log \mathcal{D}_{S|T}(S| \mathcal{E}(T))
\end{align}
and train parameters of the encoder 
($\theta_{\mathcal{E}}$) 
and both decoders 
($\theta_{\mathcal{D}}, \theta_{\mathcal{D}_{T,S}}$) 
to maximize:
\begin{align}
\argmax_{\theta_{\mathcal{E}}, \mathbin{\color{green}\theta_{\mathcal{D}}}, \mathbin{\color{orange}\theta_{\mathcal{D}_{T,S}}}}
~~
  \sum_{T, T' \in \mathcal{X}}
    \mathbin{\color{green}\mathcal{L}(T'|T)} \nonumber \\ + 
\sum_{T, S \in \mathcal{X}_A} \mathbin{\color{orange}\mathcal{L}(S|T)}
\label{eq:aux_objective}
\end{align}
Intuitively, to minimize this objective, the output of $\mathcal{E}(T)$ must
encode information about the latent situation $S$. Once encoded, this information is
accessible to the original LM text decoder $\mathcal{D}$. \cref{eq:aux_objective} is a straightforward application of standard multi-task training approaches; however, to the best of our knowledge it has not previously been used for state prediction tasks or shown to improve LMs' factual coherence.
\jda{
  Move?
  We first fine-tune the LM to convergence on $\mathcal{X}$ using
  $\mathbin{\color{green} \mathcal{L}_{T'|T}}$, then train on
  \cref{eq:aux_objective} above.
  \bzli{Appendix on training details?}
}
\subsection{Situation Prediction for Prompting}
The approach described above is general. But in LMs with very large numbers of
parameters, it might be costly to apply (or we may risk over-fitting if the fine-tuning
dataset is too small). Thus, the second approach we describe is based on
\emph{prompting} models to build better state representations.  We build on the
observation in recent work that prompts can induce models to build better task
representations by writing contents of these representations to output:
generating, then conditioning on, textual representations of useful intermediate
variables.

To induce language models to output textual state descriptions, we construct
prompts with three components: a task description $D$, a set of task
demonstrations (``training set'') $\mathcal{X}$, and an input context
$T_{pred}$.  The training set can include both unannotated and annotated
examples: unannotated examples are sequences $T_i, T'_i$, while annotated
examples are sequences $T_i, S_i, T'_i$. Formally, we construct a prompt string:
\begin{align}
\nonumber
    \mathcal{P} & = [D  \concat  \mathcal{P}_A  \concat  \mathcal{P}_U  \concat  T_{pred}] ~ , &&\textrm{where:} \\
\label{eq:aux_prompt}
    \mathcal{P}_A & = [T_0'  \concat  S_1  \concat  T_1'  \cdots   S_n  \concat  T_n' ]_{x} && \forall x \in \mathcal{X}_A \nonumber \\
    \mathcal{P}_U & = [T_0'  \concat  T_1'   \cdots   T_n' ]_{x} && \forall x \in \mathcal{X}
\end{align}
with $ \concat $ denoting string concatenation. To enable the model to \emph{predict}
annotations and text directly, each %
$S$ is prefixed with an appropriate
control token that informs the model %
that a state description string will come next.
When predicting (or scoring) a sentence $T'_\text{pred}$ in context, we first
prompt the model to generate a state representation $S_\text{pred}$, then score
$T'_\text{pred}$ conditional on $T_\text{pred}$, $S_\text{pred}$, and the entire
preceding context.
The bottom portion of \cref{fig:methods} shows a concrete example from the TRIP domain.
As above, this approach to prompting is closely related to existing ``scratchpad'' and ``chain-of-thought'' methods used for question answering and formal reasoning tasks; our auxiliary situation modeling approach applies this form of structured prompting to multi-sentence, open-ended text generation problems.

\section{Latent State Inference}
\label{sec:latent_inf}

The methods described in \cref{sec:aux_supervision} applied state supervision only to examples for which a ground-truth state
annotation was provided. For these methods to be effective, enough state
annotations must be available to provide a useful training signal in the
auxiliary loss or to specify the auxiliary prediction task for the
prompted model. But such state annotations are in general both \emph{hard
to collect} and \emph{hard to design}.

In this section we describe how to obtain them automatically,
without the need for large amounts of annotation.
Below, we re-formulate the two approaches in \cref{sec:aux_supervision} as \emph{latent variable}
models that can infer and condition on state representations even for
unannotated training documents. Intuitively, this inference problem is easier at
training time than prediction time: knowing what text followed a context
constrains the possible situations the context could describe.
Most work on semi-supervised inference of auxiliary prediction targets has
focused on automatic optimization of prompts and reasoning chains \citep{zelikman2022star,sun2022black}. %
To the best of our
knowledge, inferred latent variables have not been used to train scaffold decoders or to
design intermediate state representation for multi-step text generation. The
techniques described below are quite general, and might be productively employed
beyond the generation applications we describe here.

\subsection{Latent State Inference for Fine-Tuning}
\label{sec:latent_state_ft}

Intuitively, a good state representation is one that is both predictable from
context, and useful for predicting subsequent text. To guide inference of states for auxiliary prediction,
we introduce another encoder-decoder into the model of \cref{sec:aux_method}: one which attempts to
predict $T'$ from $S$. This model now has two pathways for predicting $T'$: one
that uses encoder representations to predict it directly from $T$, and another
which generates textual state descriptions $S$ from decoder representations,
then uses these to predict $T'$.  We train this model's parameters and infer missing
states that maximize probability of next sentences under both pathways, using
information from both $T$ and $T'$ to infer situations $S$, then using
these to directly supervise the encoder.

Formally, we optimize the complete likelihood:

\begin{align}
&\argmax_{\Theta, \hat{S}}
\sum_{\substack{T, T' \\ \in \mathcal{X}}}
    \mathbin{\color{green}\mathcal{L}(T'|T)} \nonumber \\
&\qquad~ + \sum_{\substack{T, T', S \\ \in \mathcal{X}_A}}
    \left(\mathbin{\color{orange}\mathcal{L}(S|T)} + \mathbin{\color{blue}\mathcal{L}(T'|S,T)}\right) \nonumber \\
&\qquad~ + \sum_{\substack{T, T', \hat{S} \\ \in \mathcal{X}_U}}
    \left(\mathbin{\color{orange}\mathcal{L}(\hat{S}|T)} +
    \mathbin{\color{blue}\mathcal{L}(T'|\hat{S},T)}\right) ~ .
\label{eq:EM_objective}
\end{align}

\noindent\cref{eq:EM_objective} extends auxiliary fine-tuning by concurrently training an encoder-decoder %
$\mathcal{M}_{T' \mid S,T}$
to model $\mathbin{\color{blue} p(T' \mid S,T)}$.
To optimize this objective,
we initialize $\theta_{\mathcal{E}}, \theta_{\mathcal{D}},
\theta_{\mathcal{D}_{S \mid T}}$ using \cref{eq:aux_objective}, and $\theta_{T' \mid S}$ by 
fine-tuning to convergence on %
$\mathcal{X}_A$. %
We then perform coordinate ascent (``hard EM'') by alternating between:
\begin{enumerate}
\item %
E-step: Set $\hat{S} \approx \arg\max_S ~ \mathbin{\color{orange}p(S \mid T)} \mathbin{\color{blue}p(T' \mid S)}$ for $\mathcal{X}_U$ by sampling from $\mathbin{\color{orange}p(S \mid T)}$, then reranking according to $\mathbin{\color{orange}p(S \mid T)} \mathbin{\color{blue}p(T' \mid S)}$.
\item M-step: Using the new $\hat{S}$, train $\Theta$ to maximize \cref{eq:EM_objective}. %

Rather than training to convergence, we perform SGD on \cref{eq:EM_objective} for five epochs.
\end{enumerate}
As in auxiliary fine-tuning, $\mathcal{E}$ is shared the $\mathbin{\color{green} p(T' \mid T)}$ and $\mathbin{\color{orange} p(S \mid T)}$. 
Information about inferred states shapes text generation via the auxiliary decoding objective.

\subsection{Latent State Inference for Prompting}
\label{sec:latent_state_prompt}

Work on few-shot prompting consistently finds benefits from adding extra examples to
prompts \citep{GPT3}. As in~\cref{sec:latent_state_ft}, we produce extra examples for a seed prompt by finding situation
descriptions $S$ that are predictable from $T$ and improve prediction of $T'$ on
unannotated examples. We may do so using a very similar procedure to the one in~\cref{sec:latent_state_ft}: now we choose prompts (but not model parameters) to maximize:
\begin{equation}
  \argmax_{\hat{S}} \sum_{T, T' \in \mathcal{X}_U}
  \textcolor{orange}{p(\hat{S} \mid T)} \textcolor{blue}{p(T' \mid T, S)}
\end{equation}
then add these newly annotated examples to the prompt (which we may do during
both training and evaluation).
Algorithmically, we iterate incrementally over unannotated examples
\bzl{TODO clarify}\bzl{https://aclanthology.org/N09-1069.pdf}
$\mathcal{X}_A$:
\begin{enumerate}
    \item %
    E-step: Set $\hat{S} \approx \arg\max_S \mathbin{\color{orange}p(S \mid
    T)}\mathbin{\color{blue}p(T' \mid S)}$ for each context-sentence pair
    $(T,T')$ in $\mathcal{X}_U$ by
    prompting the LM with $[D \concat \mathcal{P}_A \concat T]$, then reranking the candidate states according to
        \begin{equation} 
        \hspace{-2em}
          \textcolor{orange}{p(S \mid [D \concat \mathcal{P}_A \concat T])}
        \textcolor{blue}{p(T' \mid [D \concat \mathcal{P}_A \concat T \concat S])}
        ~ . 
        \end{equation}
    \item 
      M-step: add $[T  \concat  \hat{S}  \concat  T']$ to $\mathcal{P}_A$ in~\cref{eq:aux_prompt}.
      \jda{we haven't said anything yet about the fact that these might be
      multi-sentence / multi-state passages---should introduce earlier or not
      talk about it here}
\end{enumerate}
Once all examples in $\mathcal{X}_U$ have been annotated and added to
$\mathcal{P}_A$, we prompt with auxiliary supervision for each context in the
evaluation set using $\mathcal{P} = [D \concat \mathcal{P_A} \concat T_{pred}]$.

\section{Experimental Setup}
\label{sec:models_datasets}
\paragraph{Datasets}
We evaluate \ourmethod 
on English language modeling datasets.
\textbf{TW} is derived from %
TextWorld~\cite{ct2018textworld}.
We generate a set of 
textual game transcripts where
players navigate through a house, unlocking doors and containers to hunt for a target object. The LM is trained on these transcripts to generate next \textit{actions}. %
As state supervision, we use the set of state variables (given as %
entity-centric facts) that are \emph{known} and \emph{relevant} in the current context (see \cref{sec:choice_of_state} for more details).
\textbf{TRIP}~\cite{storks-etal-2021-tiered-reasoning} features pairs of plausible and implausible five-sentence stories which require physical commonsense reasoning to disambiguate. 
Models are trained to generate judgments of whether or not a given sentence is acceptable in a context.
The state is given by a set of attributes for each entity, which is updated after each sentence.\footnote{See~\cref{sec:app_state} for state representation details.}

Each passage $x \in \mathcal{X}$ comprises a sequence of chunks $T_0',T_1',\cdots,T_n'$. In TW,
each chunk consists of a textual action description followed by a game response.
In TRIP, each chunk is a single sentence from the story followed by a plausibility judgment.
We test coherence of generating each $T'$ from its context $T$.
For the annotated passages in $\mathcal{X}_A$, \textit{each context} $T_i$ is annotated with corresponding state information $S_i$.
Thus, passages in $\mathcal{X}_U$ can be broken down into $(T,T')$ pairs, while passages in $\mathcal{X}_A$ can be broken down into $(T,S,T')$ triples.

\paragraph{Models}
For fine-tuning experiments, we use BART-base~\cite{lewis-etal-2020-bart} %
as the language model and fine-tune it 
using the AdamW optimizer with learning rate 1e-5, stopping once validation accuracy has stopped improving for 10 epochs. %
For prompting experiments, we use the GPT3 \texttt{da-vinci-002} model~\cite{GPT3}.\footnote{Further details can be found in~\cref{sec:app_compute}.} %

\paragraph{Metrics}
To evaluate models on TW, we sample next actions from the LM and compute the fraction of these that are semantically \textbf{coherent} using the TW simulator.\footnote{In~\cref{sec:app_TW_diversity}, we also evaluate generation diversity amongst these actions using recall against the full set of ground-truth possible next actions.}
For TRIP, we evaluate every story pair by training models to predict the string \texttt{OK} or \texttt{Not OK} after each sentence depending on whether it is semantically acceptable within a given context. The TRIP dataset contains human semantic acceptability judgments for stories; we evaluate the \textbf{accuracy} with which models predict these acceptability judgments (labeling a story as unacceptable if any sentence is predicted to be unacceptable).

For TW, we report \textit{sentence-wise} metrics: we measure the fraction of next sentences which are generated to be coherent within the context. In TRIP, we report \textit{passage-wise} metrics: %
we measure the percent of complete passages for which \textit{every sentence of the passage} is labelled accurately.

\section{Experiments}
\label{sec:experiments}
\subsection{Fine-Tuning}

Our experiments use 1000 training examples, varying the fraction of these examples for which we provide state annotations
($|\mathcal{X}_A| = \{0,500,1000$ $\}$).
For each choice of $|\mathcal{X}_A|$, we repeat experiments across 8 random seeds, training on a different set of 1000 examples for each seed.
We compare models trained using ordinary language modeling techniques,
\cref{eq:aux_objective}, and \cref{eq:EM_objective}.
We evaluate using metrics described in~\cref{sec:models_datasets}.

\begin{table}[]
    \centering
    \small
    \begin{tabular}{rccccc}
    \toprule
        & $|\mathcal{X}|$ & $|\mathcal{X}_A|$ & \ourmethodtable?
        & Metric \\
    \midrule
        & & & & Coherence \\
        TW & 1k & 0 & None & $79.4\%_{\pm 2.4\%}$ \\
    \midrule
        \multirow{2}{*}{TW} & 1k & 500 & Aux Only & $80.5\%_{\pm 1.8\%}$ \\
        & 1k & 500 & Aux$+$Latent & $83.4\%_{\pm 1.4\%}$ \\
    \midrule
        TW & 1k & 1k & Aux Only & $81.5\%_{\pm 1.5\%}$ \\
    \midrule
        TW & 10k & 0 & None & $83.6\%_{\pm 2.5\%}$ \\
    \toprule
        & & & & Accuracy \\
        TRIP & 1k & 0 & None & $36.5\%_{\pm 3.5\%}$ \\
    \midrule
        \multirow{2}{*}{TRIP} & 1k & 500 & Aux Only & $43.6\%_{\pm 1.0\%}$ \\
        & 1k & 500 & Aux$+$Latent & $43.6\%_{\pm 1.0\%}$* \\
    \midrule
        TRIP & 1k & 1k & Aux Only & $43.0\%_{\pm 1.7\%}$ \\
    \bottomrule
    \end{tabular}
    \caption{BART fine-tuning results on TW and TRIP, %
    where $|\mathcal{X}|$ is the total amount of examples, and $|\mathcal{X}_A|$ is the total amount of state supervision.
    We report results for text-only querying, \ourmethod with only the auxiliary situation modeling component, and \ourmethod with both the auxiliary situation modeling and the latent situation prediction components. 
    We  report averages and standard errors over 8 random seeds.
    We see that training with any state supervision helps over training with no state supervision, and that with a comparable amount of state supervision, latent inference (sometimes) improves over auxiliary situation modeling. \\
    \small{*Latent inference unable to improve beyond base auxiliary situation modeling checkpoint}}
    \label{tab:ft_results}
\end{table}
\paragraph{Results}
Evaluation results are shown in \cref{tab:ft_results}.
In TW, 
using auxiliary supervision and latent state inference, \ourmethod with
500 state annotations improves generation coherence by \hbox{$\sim{}4\%$} over a text-only baseline, %
giving comparable improvements to
training on 9,000 more text-only examples. %
Results in~\cref{sec:app_TW_diversity} show that these improvements come at no cost to generation \emph{diversity}. %
In TRIP, %
\ourmethod with 500 seed states improves accuracy by $\sim 7\%$ over a text-only baseline. 
Note in this case that the latent inference procedure was unable to improve beyond auxiliary training.
However, even adding in the remaining 500 ground-truth state annotations does not improve the LM, %
indicating that perhaps %
the 500 seed states were sufficient for the LM to learn everything it can from state supervision.

\begin{table}[]
    \centering
    \small
    \begin{tabular}{rcccc}
    \toprule
        & $|\mathcal{X}|$ & $|\mathcal{X}_A|$ & \ourmethodtable? & Metric \\ %
    \midrule
        & & & & Coherence \\
        TW & 25 & 0 & None & $67.4\%$ \\
    \midrule
        \multirow{2}{*}{TW} & 25 & 12 & Aux Only & $68.5\%$ \\
        & 25 & 12 & Aux$+$Latent & $75.6\%$ \\
    \midrule
        TW & 25 & 25 & Aux Only & $73.9\%$ \\
    \toprule
        & & & & Accuracy \\
        TRIP & 80 & 0 & None & $59.5\%$ \\
    \midrule
        \multirow{2}{*}{TRIP} & 80 & 20 & Aux Only & $58.2\%$ \\
        & 80 & 20 & Aux$+$Latent & $67.1\%$ \\
    \midrule
        TRIP & 80 & 80 & Aux Only & $70.7\%$ \\
    \bottomrule
    \end{tabular}
    \caption{GPT3 prompting results on TW and TRIP, using text-only querying, \ourmethod with only the auxiliary situation modeling component, and \ourmethod with both the auxiliary situation modeling and the latent situation prediction components. Lines separate the amount of ground-truth manual annotation.
    We see that prompting with any state supervision helps over prompting with no state supervision, and that with a comparable amount of ground-truth state supervision, latent inference significantly improves over only auxiliary situation modeling.
    } %
    \label{tab:gpt3_results}
\end{table}

\subsection{Prompting}

In TW, we used 25 sentences (3 stories) in $\mathcal{P}$. In TRIP, we used 80 sentences (16 stories) in $\mathcal{P}$. When evaluating latent supervision, we held out state annotations on 13 sentences (2 stories) in TW, and 60 sentences (12 stories) in TRIP. 
Due to budget restrictions, we were only able to run each prompting experiment once.

\paragraph{Results}
Results are reported in~\cref{tab:gpt3_results}.
Using~\ourmethod with auxiliary situation modeling where all passages are fully annotated with state (rows 4,8) dramatically improves performance compared to a text-only baseline (rows 1,5) in both domains.
In TW, we see a $\sim 6.5\%$ improvement in generation coherence,\footnote{Results in~\cref{sec:app_TW_diversity} shows that~\ourmethod also improves generation \textit{diversity}.} while in TRIP, we see a $\sim 11\%$ improvement to accuracy of coherence judgments.

Next, we examine the setting where certain state annotations are missing from the prompt, comparing~\ourmethod with latent situation prediction (rows 3,7) against~\ourmethod with only auxiliary situation modeling (rows 2,6).
We find hat incorporating generated latent states into the prompt helps performance on both TW and TRIP, by $7.1\%$ and $8.9\%$ respectively.

\section{Analysis}
\label{sec:analysis}
\subsection{Choice of state is important}
\label{sec:choice_of_state}
\begin{table}[]
    \centering
    \begin{tabular}{rc}
    \toprule
        State Type & Coherence \\
    \midrule
        None & $79.4\%_{\pm 2.4\%}$ \\ %
        Full state & $78.0\%_{\pm 1.7\%}$ \\
        Full Known state & $79.7\%_{\pm 1.5\%}$ \\ %
        Relevant Known state & $81.5\%_{\pm 1.5\%}$ \\
    \bottomrule
    \end{tabular}
    \caption{Using %
    different subsets of the state as auxiliary supervision for TW fine-tuning yields varying amounts of coherence improvements.
    We report averages and standard errors over 4 random seeds.
    We see that the choice of state matters, and that using just the known and causally relevant portions of the state (relevant known state) outperforms using the full state.}
    \label{fig:TW_choice_of_state}
\end{table}
\begin{figure*}
    \centering
    \includegraphics[scale=0.25,trim={0 15cm 3cm 0},clip]{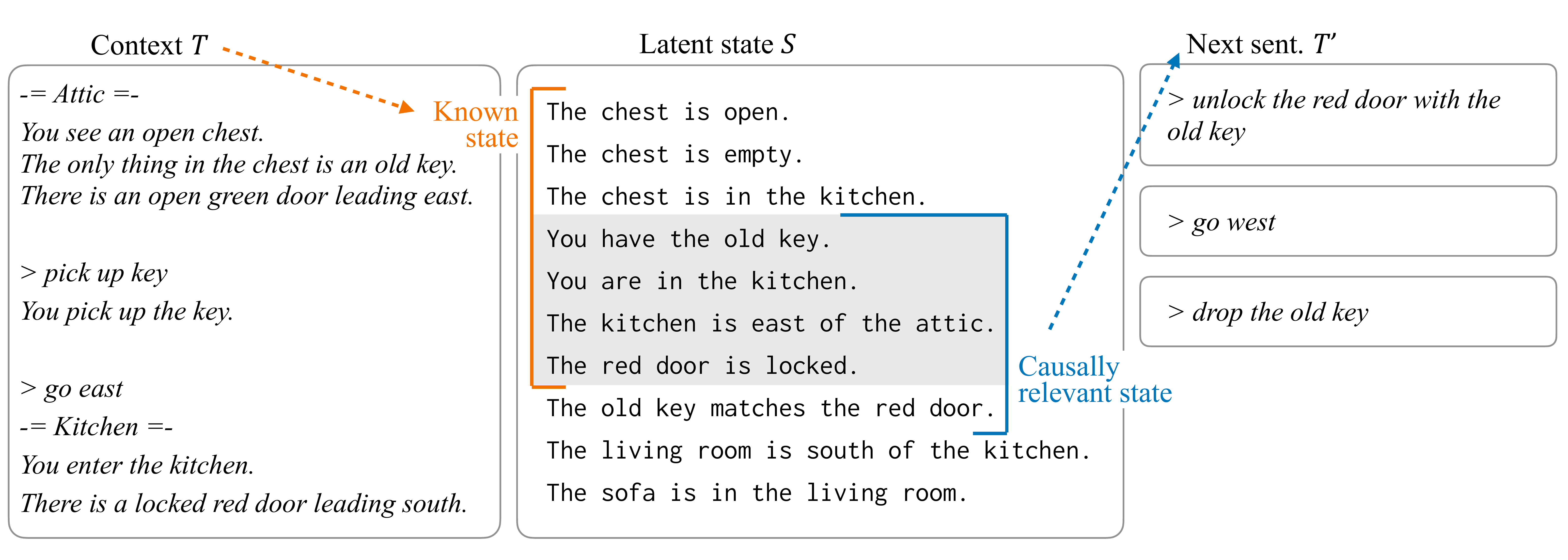}
    \caption{We find that the design of the situation representation is important: in particular, a situation representation must be ideally consist of only the intersection between the \textit{known state} and the \textit{causally relevant state} (highlighted in gray). The \textit{known state} consists of all facts deducible from the prior context $T$ (i.e. in TW only of facts about rooms or objects that the player has seen -- in this case, the kitchen and attic but not the living room). The \textit{causally relevant state} consists of all facts causally relevant to any plausible next sentence $T'$ (i.e. in TW only facts about the currently accessible items, such as, in this case, the old key, but not the chest).
    }
    \label{fig:state_choice}
\end{figure*}
\paragraph{TW} 

Because it is procedurally generated, the TW environment is able to provide detailed ground-truth state annotations for every entity that might be mentioned in text. All experiments described above use situation representations that include only a subset of entities and properties: namely (1) only those that are already \textit{known} (i.e. those have been asserted in the context), and (2) only those that are \emph{causally relevant} (i.e.\ those that, if changed, would induce a different distribution over next sentences). See~\cref{fig:state_choice}, where \textit{known facts} includes only facts deducible from prior context, while \textit{causally relevant facts} is the minimal set of facts that validate the set of plausible next sentences (e.g. no actions involving the chest are plausible next sentences, thus no facts about the chest are causally relevant). In this section, we explore the consequences of including various components of the state. %

Specifically, we train with auxiliary supervision using the three different choices of state: %
the full state, the known state (facts that satisfy condition (1)), %
and the relevant known state (facts that satisfy both conditions (1) and (2)). %
Results are shown in~\cref{fig:TW_choice_of_state}. We find that the training with the full state is %
not significantly better than simply training on text only, and %
perhaps slightly worse. Training on the subset of known facts outperforms training with the full state, and training on the intersection of known state and causally relevant state is even better.

\begin{table}[]
    \centering
    \begin{tabular}{rcccc}
    \toprule
        & $|\mathcal{X}|$ & $|\mathcal{X}_A|$ & State Type & Accuracy \\
    \midrule
        & 12 & 0 & - & $59.3\%$ \\
        TRIP & 12 & 12 & Orig & $62.8\%$ \\
        & 12 & 12 & Ours & $68.1\%$ \\
    \bottomrule
    \end{tabular}
    \caption{Using different types of state annotations for TRIP when prompting GPT3 yields various amounts of performance improvements. %
    We compare \ourmethod using TRIP's original state annotations (Orig) against \ourmethod using our own handcrafted state annotations (Ours). %
    Note that using the original state annotation is only able to improve 3.6\% over a text-only baseline, while using our state annotations improves 8.8\%.} 
    \label{tab:gpt3_state_importance}
\end{table}

\paragraph{TRIP}
Using the principles deduced from the above experiments in TW (the optimal state should be both \textit{known} from prior context and \textit{causally relevant} to the next sentence), we %
optimize the design of TRIP state annotations.\footnote{See~\cref{sec:app_state} for details.}
We used these state annotations for all experiments described above. %
In this section, we demonstrate that this outperforms using the original annotations provided in the dataset.
Specifically, we sample 12 training examples to include in the prompt,\footnote{In previous sections we used 16 samples. However, because the original states were much longer than our states, we were only able to fit 12 candidates in context using the original state annotations.} and compare text-only prompting against \ourmethod with the original states (Orig) and \ourmethod with handcrafted states (Ours). %
Results are reported in~\cref{tab:gpt3_state_importance}. By using our handcrafted states, we were able to achieve a much higher accuracy than using the original states.

\subsection{Sampling and reranking prompt states are better than greedily generating best states}
\begin{table}[]
    \centering
    \begin{tabular}{rcccc}
    \toprule
         & TW & TRIP \\
    \midrule
        \ourmethod & 75.6\% & 67.1\% \\
         without state reranking & 72.4\% & 65.8\% \\
    \bottomrule
    \end{tabular}
    \caption{Ablating state reranking with $p(T'\mid S)$ when inferring the optimal latent state for prompting. In both TW and TRIP, \ourmethod works better when we sample multiple states from $p(S\mid T)$ and rerank according to $p(T' \mid S)$, than when we simply take the greedy optimal state from $p(S\mid T)$.}%
    \label{tab:state_generation}
\end{table}
A simplification of our latent state inference procedure for prompting simply asks GPT3 to greedily generate the most likely state according to prior context (i.e., $\argmax_S p(S \mid T)$), without considering $p(T' \mid S)$. %
We compare our currently latent state procedure against this greedy state generation baseline in~\cref{tab:state_generation}. We find that it indeed helps to consider $p(T' \mid S)$ when generating states, improving next sentence coherence by 3.2\% in TW and next sentence accuracy by 1.3\% in TRIP.

\subsection{For a fixed context window budget, including more state annotations outperforms including more text samples}
\label{sec:fixed_ctxt_budget}
\begin{table}[]
    \centering
    \begin{tabular}{ccccc}
    \toprule
        & \# toks & $|\mathcal{X}|$ & $|\mathcal{X}_A|$ & Metric \\
    \midrule
        & & & & Coherence \\
        TW & 3199 & 54 & 0 & 56.7\%* \\
        TW & 3199 & 25 & 25 & 65.0\%* \\
    \midrule
        & & & & Accuracy \\
        TRIP & 3053 & 229 & 0 & 60.5\% \\
        TRIP & 3054 & 80 & 80 & 70.7\% \\
    \bottomrule
    \end{tabular}
    \caption{When prompting with limited context-window size, supplementing existing prompt demonstrations with states is more token-efficient than providing more text-only training examples. \\
    \small{*Coherences of \textit{greedy} next generations are reported in this experiment for TW.}
    }
    \label{tab:fixed_ctxt_budget}
\end{table}

Because the limiting factor in many current few-shot prompting methods is context window size rather than annotation effort,
we study whether it is more token-efficient
to include additional state annotations or additional text examples in the prompt.
We compute the number of tokens in prompts annotated with state ($\mathcal{P}_A$), then formulate a text-only prompt ($\mathcal{P}_T$) by
stripping the state annotations from $\mathcal{P}_A$, then appending randomly-selected text-only samples from the remaining training data until the number of tokens in the new prompt is equal (or nearly equal) to the number of tokens in $\mathcal{P}_A$.

We prompt the LM using either text-only prompting conditioned on $\mathcal{P}_T$, or auxiliary prompting conditioned on $\mathcal{P}_A$. The results are shown in~\cref{tab:fixed_ctxt_budget}. (Due to limitations in annotation budget, for TW in this experiment, we report coherence of the greedily-generated next actions rather than sampling 5 actions.) We see that under a fixed context token budget, in both domains, it is more helpful to supplement existing examples with their state annotations rather than insert additional text-only examples into the context window.

\section{Conclusion}
Effective generation of coherent text requires reasoning about the world that text describes.
In this work, we use entity states as auxiliary supervision to improve LMs ability to perform this reasoning under both fine-tuning and prompting.
We find that when either annotation budget (for fine-tuning) or context window size (for prompting) are limited, it is more sample- and token-efficient to increase the amount of state supervision rather than text-only supervision.
However, since state annotations are harder to collect, %
we introduce latent supervision algorithms %
for \textit{sample-efficiently} improving LM generation coherence, and %
demonstrate improvements in two domains.
Our results point to a potentially broad role for semantic supervision in LM training and prompting---even small amounts can yield large coherence improvements.
This work more broadly suggests that semantic state reasoning is still challenging for even modern large language models, and but can be improved without fundamental changes to the architecture of existing LMs.

\bibliography{anthology,custom}
\bibliographystyle{acl_natbib}

\newpage
\appendix
\label{sec:appendix}

\section{Constructing the State}
\label{sec:app_state}
In each domain, the state is a collection of facts (attributes and/or relations) about each entity.
It is updated each time there is a %
new action, instruction, or sentence.
We convert the state to natural language to take advantage of existing linguistic understanding in pre-trained models. %
Future work can examine the effect of using non-natural-language forms of state.

Below, we discuss the details of this conversion from the available state annotations in each domains.

\paragraph{TW}
In TW, the simulator gives us the \textbf{full state}, or the full set of facts describing the state of the world after executing each agent action. Facts are either entity properties (e.g. \texttt{locked(door)}), or relations between two entities (e.g. \texttt{is-in(key, chest)}).
However, since the agent has not explored the full state at the start of each game, at each step, we compute a subset of the facts that the agent \textit{knows about}. We call this the \textbf{known state}.
We further restrict this subset to only facts that are \textit{causally relevant} to any possible next action that the agent can take, such that all possible next actions can be inferred from just this set. We call this the \textbf{relevant known state}. %

We compute both these sets heuristically: the known state consists of all facts about any currently or previously accessible entities that the agent has encountered. %
For the \textit{relevant} known state, we discard facts about previously accessible entities and keep only facts about currently accessible entities. Specifically, the relevant known state consists of facts about: 1. player location, 2. all currently accessible items (i.e. in the current room or in the inventory), 3. which doorways are accessible from the current room and/or which rooms neighbor the current room.

We convert %
collections of facts to natural language following the same procedure as~\newcite{li-etal-2021-implicit}. Specifically, propositions $p(o)$ are converted to ``\textit{the $\{o\}$ is $\{p\}$}'', while relations $r(o_1,o_2)$ are converted to ``\textit{the $\{o_1\}$ is $\{r\}$ $\{o_2\}$}''.

\paragraph{TRIP}
In TRIP, we write out seed states for 16 stories, consisting of facts known to hold true after each sentence of the story --- then use GPT3 to automatically infer states for the remaining stories in the training data.
We aim to construct the state in TRIP to capture the spirit of the \textbf{relevant known state} in TW (which we know from~\cref{sec:choice_of_state} to be the optimal state supervision), whereby we only include facts both known from the prior context and potentially causally relevant to the next sentence.
However, though capturing known facts is straightforward, because TRIP is a real dataset consisting of open-ended text, the set of plausible next generations is open-ended, meaning that the full set of causally relevant known facts cannot be always be anticipated ahead of time. %
Instead, we use the ground-truth acceptable completion as a minimal guarantee
-- we aim to include facts informative for generating at least the single ground-truth next sentence in the acceptable story (which isn't always straightforwardly derived from the known facts). One example is as follows:
\begin{itemize}
\item $T = $ \textit{Tom packed his gloves in his suitcase. Tom checked his suitcase in at the airport.}
\item $S = $ \texttt{Tom's gloves are in the suitcase. The suitcase is checked in at the airport. Tom does not have his suitcase. Tom does not have his gloves.}
\item $T' = $ \textit{Tom boarded the plane without his gloves.}
\end{itemize}

Note that while \textit{Tom does not have his gloves} is technically inferrable from \texttt{Tom's gloves are in the suitcase. The suitcase is checked in at the airport}, including this fact explicitly in $S$ reinforces the causal link between the next sentence $T'$ and $S$.

For the analysis in~\cref{sec:choice_of_state}, we compare against a stringified version of the originally-provided states.
In the original dataset, each sentence of a story is annotated with the %
state changes applied to each of the (up to 15) attributes of that entity. The state annotations take the form of $(\textit{entity}, \textit{attribute}, \textit{value})$ triples.
Each entity attribute is %
associated
with a value indicating the direction of change for that attribute. For example, $(\textit{shirt}, \textit{cleanliness}, \textit{true}\to\textit{false})$ indicates \textit{the shirt became dirty}.

Because there are a finite set of (15) attributes and (8) values, we enumerate rules for converting all $(\textit{attribute}, \textit{value})$ pairs to natural language predicates \texttt{VP}. We then convert $(\textit{entity}, \textit{attribute}, \textit{value})$ triples into ``\textit{the $\{$entity$\}$} \texttt{VP}''.

\section{Effect of \ourmethod on Generation Diversity in TW}
\label{sec:app_TW_diversity}
\begin{table}[]
    \small
    \centering
    \begin{tabular}{rcccc}
    \toprule
        & $|\mathcal{X}|$ & $|\mathcal{X}_A|$ & \ourmethodtable? & Recall \\ %
    \midrule
        FT & 1k & 0 & None & $11.8\%_{\pm 0.3\%}$ \\
    \midrule
        \multirow{2}{*}{FT} & 1k & 500 & Aux Only & $11.8\%_{\pm 0.3\%}$ \\
        & 1k & 500 & Aux$+$Latent & $11.9\%_{\pm 0.2\%}$ \\
    \midrule
        FT & 1k & 1k & Aux Only & $12.6\%_{\pm 0.4\%}$ \\
    \toprule
        Prompt & 25 & 0 & None & $33.3\%$ \\
    \midrule
        \multirow{2}{*}{Prompt} & 25 & 12 & Aux Only & $40.1\%$ \\
        & 25 & 12 & Aux$+$Latent & $42.1 \%$ \\
    \midrule
        Prompt & 25 & 25 & Aux Only & $40.9\%$ \\
    \bottomrule
    \end{tabular}
    \caption{TW generation diversity for fine-tuning and prompting with and without components of~\ourmethod. We see that using~\ourmethod with fine-tuning does not harm the diversity of generated samples, while using~\ourmethod with prompting actually increases diversity.}
    \label{tab:TW_state_diversity}
\end{table}

To measure the diversity of LM outputs, we use \textit{recall}\footnote{Because we sample at most 5 unique generations from the LM, there is a hard ceiling on maximum achievable ``recall'' in our case.} between the set of LM generations and the full set of ground-truth valid sentences. This latter set is provided to us by the TextWorld simulator.
Note that this set is not entirely complete, as there will be generations that are consistent with the \textit{known facts} from the prior context but contradict an \textit{unknown fact}, and is consequently not accepted by the simulator.
However, recall against the simulator-provided set of valid sentences remains a good heuristic for diversity.

We examine how training with~\ourmethod affects generation diversity. 
We use the same models and training/prompting setups as in \cref{sec:experiments} %
and evaluate the diversity among the generated samples.
Results are shown in~\cref{tab:TW_state_diversity}.
We showed in \cref{sec:experiments} that \ourmethod improves TW generation coherence in both the fine-tuning and prompting cases.
As shown in~\cref{tab:TW_state_diversity},
\ourmethod does not sacrifices 
diversity to achieve those coherence gains. %
In fact, prompting with \ourmethod \textit{improves} diversity when compared against a text-only model, and doing latent inference appears to additionally improve diversity beyond simply auxiliary situation modeling.

\section{Infrastructure and Reproducibility}
\label{sec:app_compute}
We ran all fine-tuning experiments on a single 32GB NVIDIA Tesla V100 GPU. We use a BART-base model which has 6 Transformer layers each in its encoder and decoder, and 139M total parameters. Training time varies depending on domain and data size, but generally is not longer than a few hours. As a reference point: on 1000 TW examples, training takes $\sim$1 hour for text-only training, $\sim$1-2 hours for training with auxiliary state supervision, and $\sim$1-3 hours for training with latent state supervision.
For prompting results, we use OpenAI's GPT3 \texttt{text-davinci-002} model. For sampling next actions in TW, we use a generation temperature of 0.7. When judging acceptability of each sentence in TRIP, we directly compare $p(\text{Not OK})$ against $p(\text{OK})$. 
When sampling states for latent state inference, to encourage diversity, we use a generation temperature of 0.9.

\end{document}